%% file: main.tex
\documentclass[letterpaper, 10 pt, conference]{ieeeconf}

\IEEEoverridecommandlockouts                               
\usepackage{graphicx}
\usepackage{amsmath}
\usepackage{amssymb}
\usepackage{subfig}
\usepackage{float}
\usepackage{titlesec}
\usepackage{mathtools}
\usepackage[nolist]{acronym}

\usepackage{booktabs}
\usepackage{multirow}
\usepackage{tikz}

\makeatletter
\let\NAT@parse\undefined
\makeatother

\usepackage[breaklinks=true,colorlinks,bookmarks=false]{hyperref}
\usepackage[noabbrev,capitalise]{cleveref}

\usepackage{url}
\usepackage{color}

\titlespacing*{\section}{0pt}{1.2mm}{1.2mm}
\titlespacing*{\subsection}{0pt}{.7mm}{.7mm}
\titlespacing*{\subsubsection}{0pt}{1mm}{1mm}

\input{notation.tex}

\renewcommand{\baselinestretch}{0.935}

\title{\LARGE \bf
Leveraging Neural Network Gradients within Trajectory Optimization for Proactive Human-Robot Interactions
}

\author{Simon Schaefer$^1$, Karen Leung$^2$, Boris Ivanovic$^2$, Marco Pavone$^2$
\thanks{This work was supported by the Werner Siemens Scholarship (WSS), Office of Naval Research (Grant N00014- 17-1-2433), and Toyota Research Institute (“TRI”). This article solely reflects the opinions and conclusions of its authors and not WSS, ONR, TRI, or any other Toyota entity.}
\thanks{We thank Amine Elhafsi and Tim Salzmann for their useful feedback and discussions.}%
\thanks{$^1$Institute of Dynamic Systems and Control, ETH Zurich, Zurich,
Switzerland. {\tt\small  sischaef@student.ethz.ch}}
\thanks{$^2$Department of Aeronautics and Astronautics, Stanford University, USA. {\tt\small  \{karenl7, borisi, pavone\}@stanford.edu}}
}

\begin{document}

\input{acronyms}

\maketitle
\thispagestyle{empty}
\pagestyle{empty}

\begin{abstract}
To achieve seamless human-robot interactions, robots need to intimately reason about complex interaction dynamics and future human behaviors within their motion planning process.
However, there is a disconnect between state-of-the-art neural network-based human behavior models and robot motion planners---either the behavior models are limited in their consideration of downstream planning or a simplified behavior model is used to ensure tractability of the planning problem.
In this work, we present a framework that fuses together the interpretability and flexibility of trajectory optimization (TO)
with the predictive power of state-of-the-art human trajectory prediction models.
In particular, we leverage gradient information from data-driven prediction models to explicitly reason about human-robot interaction dynamics within a gradient-based TO problem.
We demonstrate the efficacy of our approach in a multi-agent scenario whereby a robot is required to safely and efficiently navigate through a crowd of up to ten pedestrians.
We compare against a variety of planning methods, and show that by explicitly accounting for interaction dynamics within the planner, our method offers safer and more efficient behaviors, even yielding proactive and nuanced behaviors such as waiting for a pedestrian to pass before moving.
\end{abstract}

\section{Introduction}\label{sec:intro}
Robots that operate alongside humans, such as autonomous cars and delivery robots, need to proactively plan in highly dynamic and stochastic environments characterized primarily by the uncertainty stemming from human behaviors.
In such settings, a robot must reason about how its future actions may affect the behavior of those around it and plan accordingly. Furthermore, in settings with time-sensitive agents (e.g., humans working in a hospital), we desire robots to impact surrounding humans as little as possible. 

Observing data from human-human interactions has provided valuable insight into modeling human behaviors and interaction dynamics, and such data-driven models are beginning to play a larger role within a robot's 
decision-making process \cite{RudenkoPalmieriEtAl2019}.
A robot may reason about human actions, and corresponding likelihoods, based on how it has seen humans behave in similar settings.
To implement a robot's policy, model-free methods tackle this problem in an end-to-end fashion---human behavior predictions are implicitly encoded in the robot's policy which is learned directly from data. However, this often leads to opaque solutions, thus limiting the interpretability of a robot's decision-making process.

On the other hand, model-based methods decouple model learning and policy construction---a probabilistic understanding of the interaction dynamics is used as a basis for policy construction.
By decoupling action/reaction prediction from policy construction, model-based approaches (i) afford a degree of transparency in a planner’s decision making that is typically unavailable in model-free approaches, (ii) are able to alter the policy to induce a desired behavior at run time, and (iii) can be applied to a variety of settings by varying elements of the model and/or policy whereas model-free methods would need to be retrained from scratch.

In this paper, we direct our focus towards model-based approaches and propose a flexible computational paradigm that enables the use of data-driven prediction models within gradient-based \ac{TO} problems. In particular, our framework is highly relevant to robot motion planning for human-robot interactions as it brings together the expressive powers of deep neural networks for human behavior prediction with the transparency of \ac{TO}.

\begin{figure}
    \centering
    \includegraphics[width=0.45\textwidth]{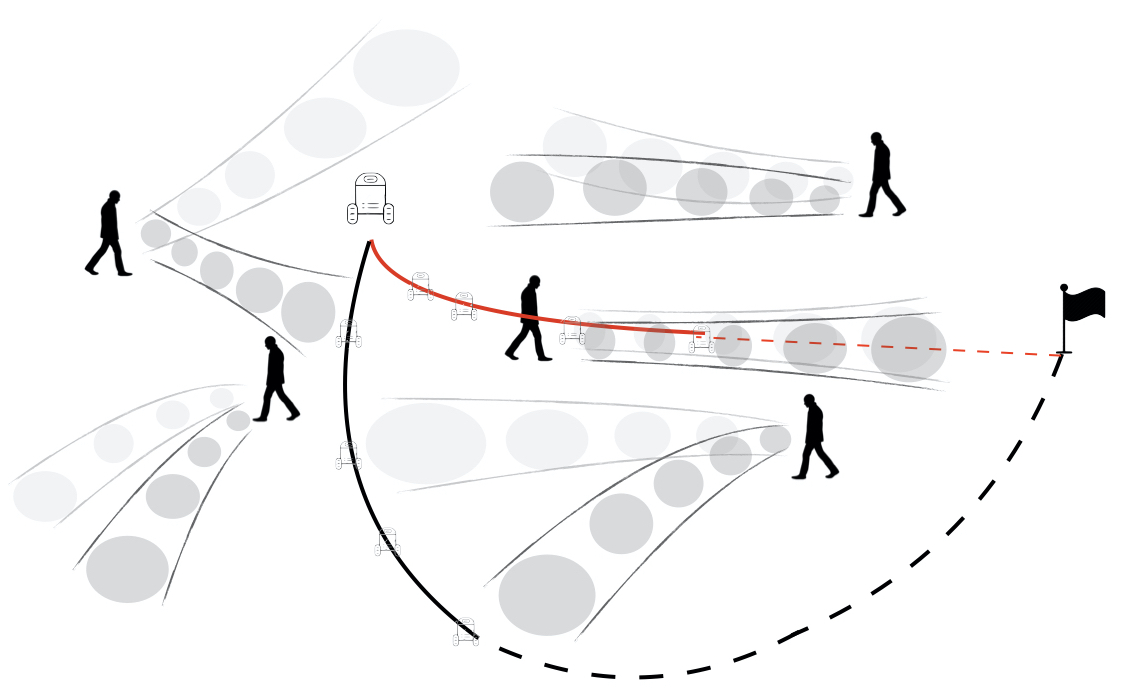}
    \caption{Our robot planning algorithm utilizes gradients from deep generative human trajectory prediction models. The robot waits before moving directly to the goal (in red) instead of naively planning a large and inefficient detour (in black).}
    \label{fig:hero}
\end{figure}

\noindent \textbf{Contributions:} 
Motivated by the problem of safe and interactive planning for human-robot interactions, we propose a novel motion planning framework that incorporates the gradient information of deep neural networks within a nonlinear \ac{TO} problem to produce safe and proactive robot plans that take into account possible human reactions.
Our contributions are two-fold: 
(i) We present a method for explicitly incorporating prediction model gradients in gradient-based optimization solvers, and show that it yields proactive and nuanced robot behaviors.
(ii) Utilizing (i), we present a socially-aware robot navigation framework that leverages multimodal probabilistic human trajectory predictions to produce \emph{safe} and \emph{minimally-invasive} robot plans, i.e., non-colliding plans that minimally interfere with humans' natural motion.
We demonstrate the efficacy of our approach with experiments involving a robot navigating through a crowd of simulated pedestrians. 
We compare against other model-based planning paradigms and show that our approach achieves better performance across several metrics.

\section{Related Work}\label{sec:related_work}
We review relevant work in \ac{TO} in Section~\ref{subsec:related trajectory opt}, discuss human behavior prediction models and their usage within planning frameworks in Section~\ref{subsec:related behavior prediction}, and end with a discussion on safety in Section~\ref{subsec:related safe planning}.

\subsection{Trajectory Optimization}\label{subsec:related trajectory opt}
Robot motion planning under uncertainty can be formalized as a \ac{POMDP} \cite{KaelblingLittmanEtAl1998}, where one seeks to design a policy that maps state probability distributions to actions.
Despite the theoretical and practical successes of \ac{POMDP} theory \cite{KurniawatiHsuEtAl2008,SomaniYeEtAl2013,KurniawatiDuEtAl2010,SunbergKochenderfer2018,BaiCaiEtAl2015}, online computation of a robot policy is extremely computationally intensive, motivating the need to consider open-loop planning.
By replanning frequently, the robot is able to update its open-loop plans to adjust for observed changes in the environment, such as a moving obstacle. 
\ac{TO}, specifically direct methods, is a popular open-loop method that solves for a sequence of states and controls that minimizes a planning objective, such as travel time, subject to dynamic, state, and control constraints. 
There are many modern direct \ac{TO} methods such as \cite{KalakrishnanChittaEtAl2011,ZuckerRatliffEtAl2013,SchulmanDuanEtAl2014} that can compute collision-free trajectories for complex high-dimensional robotic systems, e.g., a 34-dimensional humanoid robot.
Advantages of using \ac{TO} include the transparency and interpretability of the algorithm, ability to leverage highly-optimized off-the-shelf solvers, adaptability to new environments, and flexibility for a designer to specify desired behaviors at run-time.
However, a majority of \ac{TO} techniques consider a \emph{static} environment and rely on fast replanning to account for moving obstacles, which may produce naive results when interacting with intelligent agents (i.e., humans).
When dealing with uncertainty, many \ac{TO} problems are limited to Gaussian assumptions on the robot's dynamics \cite{LewBonalliECC2020,OzakiCampagnolaEtAl2020}.
As such, designing tractable robot \ac{TO} algorithms that reason about complex uncertainties and dynamic interactions with the environment, such as multimodal uncertainties stemming from human-robot interactions, remains a challenge. 

\subsection{Data-driven Human Behavior Prediction Models}\label{subsec:related behavior prediction}

Recent advances in deep neural networks and the availability of large datasets have enabled highly expressive human behavior prediction models which can learn arbitrarily complex (i.e., multimodal) probability distributions conditioned on many types of variables such as past interactions, images, maps, agent-type, etc. (e.g., \cite{SalzmannIvanovicEtAl2020,GuptaJohnsonEtAl2018,KosarajuSadeghianEtAl2019}). Despite their success in producing highly-accurate human behavior predictions,
a majority of these models remain disconnected from downstream planning and control. Only recently have behavior prediction models been designed that consider downstream planning algorithms \cite{IvanovicElhafsiEtAl2020}, creating synergy between the prediction model and planner.
Since many existing deep learned prediction models produce samples of future human trajectories, a robot planner may rely on search-based planning algorithms to incorporate this type of output structure  \cite{SchmerlingLeungEtAl2018,NishimuraIvanovicEtAl2020}. While search-based methods show promising results, the quality of the search is limited by the replanning rate thereby potentially missing rare yet catastrophic outcomes. Further, nuanced behaviors may be precluded by the discrete nature of the search and safety is only incentivized via the objective function rather than treated as a constraint.
As such, developing synergistic robot planning algorithms that can fully utilize highly-expressive deep learned human behavior prediction models yet also incorporate important planning constraints, such as safety, remains a challenge in this field.

In contrast to deep human behavior prediction models, \cite{SadighSastryEtAl2016c} uses handcrafted interpretable features to learn an analytic representation of a human's internal reward function with Inverse Reinforcement Learning \cite{ZiebartMaasEtAl2008,LevineKoltun2012}. In \cite{SadighSastryEtAl2016c}, gradient information from the human behavior model was directly incorporated into a \ac{TO} problem and an off-the-shelf gradient-based solver was used to solve it. By tightly coupling modeling and control, the robot was able to effectively leverage the human behavior model within the optimization problem and produce human-interpretable behaviors without relying on hand-coded heuristics. We strive to follow this paradigm of using gradient information from the human behavior prediction model directly in the \ac{TO} problem. In particular, we aim to develop a paradigm that enables the use of complex deep neural network models within a \ac{TO} formulation since many of these models are state-of-the-art in human behavior prediction (see the nuScenes \cite{CaesarBankitiEtAl2019} challenge).

\subsection{Safe Planning in Stochastic Environments}\label{subsec:related safe planning}
Ensuring safety for human-robot interactions is paramount, yet challenging due to the uncertainty in human behaviors, especially since interactions inherently require the robot and humans to operate in close-proximity with each other.
Safety can be incentivized via the planning cost \cite{SchmerlingLeungEtAl2018, SadighSastryEtAl2016c}, but it competes with other planning objectives and thus is not treated rigorously. Alternatively, safety can be enforced by only selecting plans that avoid the forward reachable set of other agents \cite{AlthoffDolan2011,LorenzettiChenEtAl2018} or planning robust trajectory tubes that provide a buffer between the robot and obstacles \cite{MajumdarTedrake2017,SinghMajumdarEtAl2017,Fridovich-KeilHerbertEtAl2018}. However, many of these methods account for uncertainty in the robot's dynamics and/or static obstacles, and not for the coupling between the robot and humans' dynamics. Inspired by \cite{LeungSchmerlingEtAl2019}, we leverage \ac{HJ} backward reachability analysis \cite{MitchellBayenEtAl2005}, a formal verification tool for guaranteeing closed-loop safety of a system subjected to disturbances from the environment. In particular, safety is treated as a control constraint that can easily be applied to many \ac{TO} formulations. Notably, \ac{HJ} reachability is independent of the prediction or planning model used, making it a useful and flexible tool in ensuring safety for a wide range of applications (e.g., \cite{FisacAkametaluEtAl2018,LeungSchmerlingEtAl2019,BajcsyBansalEtAl2019,WangLeungEtAl2020}).

In this work, we present a direct \ac{TO} method for robot motion planning that
leverages state-of-the-art deep human behavior prediction models to produce safe and proactive robot behaviors. Since gradient information from deep neural networks is readily available, we propose using such models within gradient-based \ac{TO} to enable wider usage of state-of-the-art human behavior prediction models for proactive robot planning.

\section{Problem Formulation}\label{sec:problem_formulation}
Let $\xrobt{t} \in \mathcal{X}_\mathrm{R} \subset  \mathbb{R}^{n_\mathrm{R}}$ and $\urobt{t} \in \mathcal{U}_\mathrm{R} \subset \mathbb{R}^{m_\mathrm{R}}$ be a robot's state and control at time $t$ respectively. Let $N$ be the number of humans in the environment, and $\xhumt{t} \in \mathcal{X}_\mathrm{H} \subset \mathbb{R}^{Nn_\mathrm{H}}$ and $\uhumt{t} \in \mathcal{U}_\mathrm{H} \subset \mathbb{R}^{Nm_\mathrm{H}}$ be the humans' state and control at time $t$. For ease of notation, we sometimes use $\uhum^k$ to indicate the $k$th human's control, and $\xrobt{a:b}$ denotes a sequence of robot states from time step $a$ to time step $b$ (this applies to controls, and when describing humans too).
Let the deterministic, time-invariant, discrete-time state space dynamics of a robot and humans be given by $\xrobt{t+1} = f_\mathrm{R}(\xrobt{t}, \urobt{t})$ and $\xhumt{t+1} = f_\mathrm{H}(\xhumt{t}, \uhumt{t})$ respectively.
 Since the robot and humans are interacting with each other, their controls are coupled---we assume that a human's next control depends on the robot's next control. Specifically, we assume that at each time step $t$, the humans' next control is drawn from a distribution conditioned on the joint interaction history ranging from the initial time-step $\tau = 0$ to $\tau = t$, $\historyt{0:t} = (\xrobt{0:t}, \urobt{0:t}, \xhumt{0:t}, \uhumt{0:t})$\footnote{For ease of notation, we drop the subscript on $h$ to indicate the entire interaction history up till the current time $t$ which should be known from context.}, and the robot's next control $\urobt{t+1}$. That is, $\Uhumt{t+1} \sim P(\historyt{0:t}, \urobt{t+1})$ is a random variable (capitalized to distinguish from a drawn value $\uhumt{t+1}$). We can iteratively propagate the humans' dynamics and sample from $P(\historyt{0:t}, \urobt{t+1})$ over a time horizon $T$, thus a robot may reason about the random variable,
\begin{equation}\label{eq:human_action_prob}
    \Uhumt{t+1:t+T} \sim P(\historyt{0:t}, \urobt{t+1:t+T}),
\end{equation}
the humans' future control sequence in response to the robot's future control sequence $\urobt{t+1:t+T}$.

The goal of a robot interaction planner is to select a control sequence $\urobt{t+1:t+T}$ that minimizes a planning cost $J$ that depends on the robot's future interaction with the humans subject to state, control, and dynamics constraints. 
Due to the stochasticity of human behavior and the complex coupling between the robot and humans' controls, $J$ is not deterministic because it depends on $\uhumt{t+1:t+T}$, samples from the distribution in \eqref{eq:human_action_prob}. 
We describe how to incorporate this uncertainty into a \ac{TO} problem in Section~\ref{subsec:interactive_loss}.
Additionally, as safety is an extremely important consideration, we treat safety as a \emph{constraint} in our \ac{TO} problem via \ac{HJ} reachability as discussed in Section~\ref{subsec:related safe planning}, with details given in Section~\ref{subsec:safety_constraints}.
Finally, we solve the described optimization problem in a receding horizon fashion, executing the first control $\urobt{t+1}$ before replanning again. The final formulation for this \ac{TO} problem is presented in Section~\ref{subsec:trajopt}.

\section{Proactive Trajectory Optimization for Human-Robot Interactions}\label{sec:mantrap}
In this section, we describe how to tackle the interactive trajectory optimization problem as described in Section \ref{sec:problem_formulation} in the case where the humans' future controls are predicted with a neural network model. 
We consider a multi-agent human-robot interaction: a robot must navigate past multiple humans to reach a goal state whilst avoiding collisions, similar to the setting illustrated in Figure \ref{fig:hero}.
\subsection{Multimodal Human Trajectory Prediction Model}\label{subsec:prediction_model}
A challenging component of human-robot interactions is the multimodal uncertainty present in human behavior. As discussed in Section \ref{subsec:related behavior prediction}, neural networks dominate state-of-the-art methods for human behavior prediction. In this work, we consider Trajectron++\footnote{We use Trajectron++ trained on the ETH dataset discussed in \cite{PellegriniEssEtAl2009}.} \cite{SalzmannIvanovicEtAl2020} to model \eqref{eq:human_action_prob}. Trajectron++ is a recurrent \ac{CVAE}, a latent variable model, that produces a multimodal distribution over future human control sequences conditioned on interaction history and future robot control sequences.
Each mode of the multimodal output distribution is induced by conditioning on a discrete latent variable $z$; for each value of $z$, the model produces a sequence of normal distributions $\lbrace\mathcal{N}(\mu_{\tau\mid z}, \Sigma_{\tau\mid z})\rbrace_{\tau=0,...,T}$ over the humans' future control sequence. 
Figure \ref{fig:hero} illustrates the predictions that Trajectron++ can produce: the ellipses represent the normal distributions over position at each time step, and the trajectory ``tubes'' emanating from each human correspond to trajectory distributions conditioned on different latent values.
See \cite{IvanovicLeungEtAl2020} for a self-contained tutorial on recurrent \ac{CVAE}-based human trajectory prediction models, including Trajectron++.

Our method is also applicable to a wider class of prediction models as long as (i) the output predictions depend on the robot's future controls, and (ii) the model is differentiable with respect to future robot controls. Though not strictly necessary, the outputs should represent distribution parameters instead of samples. While it is possible to fit a distribution over samples, this adds an additional layer of computation and a potential reduction in prediction accuracy.

\subsection{Interactive Loss Function}\label{subsec:interactive_loss}

\begin{figure}
    \centering
    \includegraphics[width=0.48\textwidth]{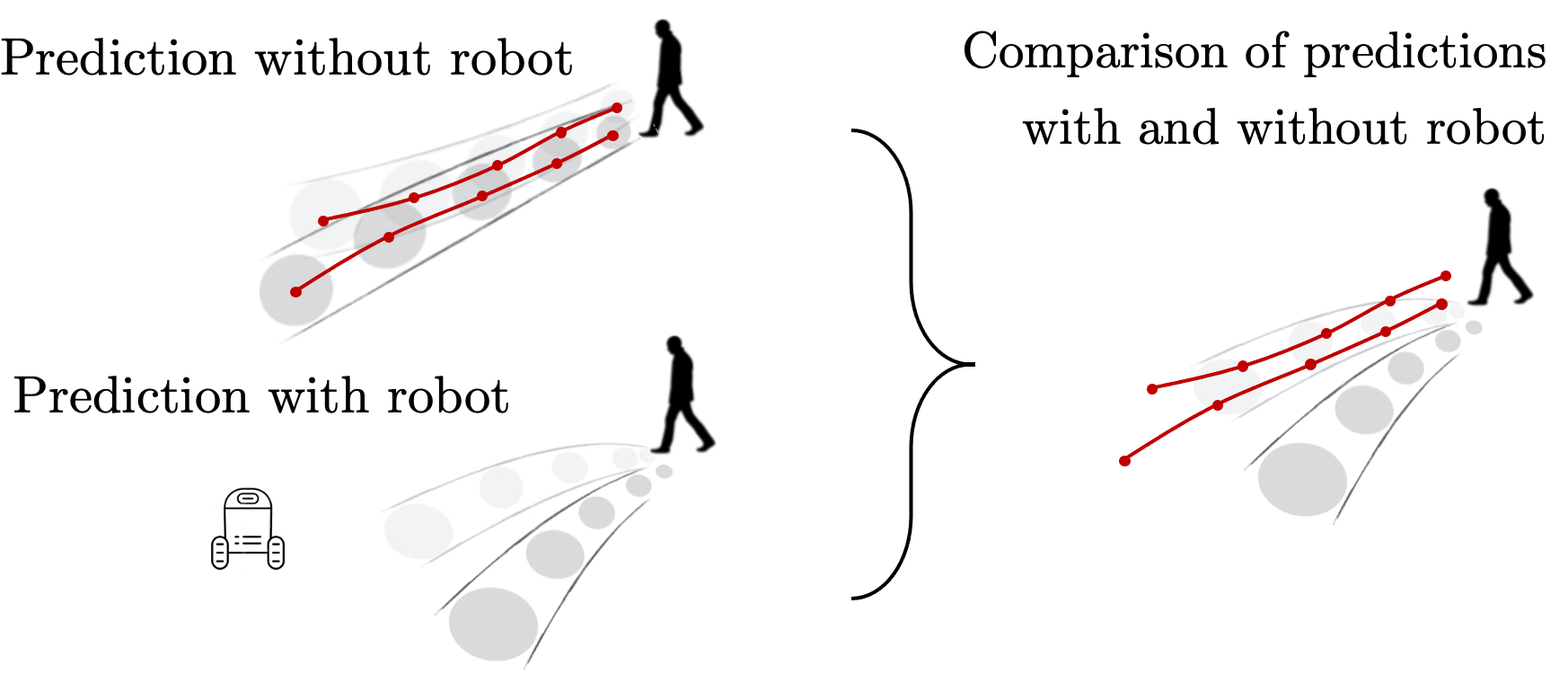}
    \caption{To measure the robot's invasiveness on human $k$'s motion, we compute the mean of each trajectory mode (red points and lines) of $\puncond{k}$ (top left) and compute the likelihood of each mean (right) under the distribution $\pcond{k}$ (bottom left).}
    \label{fig:minimally invasive}
\end{figure}

We analyze the invasiveness of the robot's plan by comparing the predictions of human behavior when conditioned and unconditioned on the robot's plan.
In short, we aim to select robot controls such that the humans behave as if the robot were not there. 
As a result, the robot is discouraged from selecting plans that interrupt the flow of a human's motion or is very close to humans.
Let $\puncond{k}$ represent the (multimodal) \ac{pdf} of human $k$'s future trajectory that \emph{is not} conditioned on robot future controls\footnote{Depending on the prediction model used, there are various ways that $\puncond{k}$ can be obtained. For Trajectron++, we place zero attention on the robot when computing agent-agent influences, therefore removing the robot's influence in the scene.}, and $\pcond{k}$ be the (multimodal) \ac{pdf} of human $k$'s future trajectory that \emph{is} conditioned on robot future controls.
Ideally, we would like to minimize the statistical distance between $\puncond{k}$ and $\pcond{k}$, e.g., KL-divergence. However, since these distributions are multimodal, computing the divergence is challenging and expensive. As a simplification, for each human $k$, we strive to maximize the likelihood of seeing the mean of each mode of $\puncond{k}$ with respect to $\pcond{k}$. That is, for $N$ pedestrians and $Z$ modes, the interaction cost $J_\mathrm{int}$ at time $t$ that we want to \emph{minimize} is,

\vspace{-3mm}
{\small
\begin{equation}
    J_\mathrm{int}(\urobt{t+1:t+T}; h) = -\sum_{k=1}^N \sum_{i=1}^{Z}\log{ \pcond{k}(\bar{u}_{\mathrm{H}, t+1:t+T \mid z_i}^k)},
\end{equation}}
\normalsize

where $\bar{u}_{\mathrm{H}, t+1:t+T \mid z}^k = \mathbb{E}_{\puncond{k}(\uhumt{t+1:t+T}^k \mid z)}[\uhumt{t+1:t+T}^k]$.
In essence, given \eqref{eq:human_action_prob}, $J_\mathrm{int}$ penalizes the difference between $\pcond{k}$ and $\puncond{k}$, illustrated in \cref{fig:minimally invasive} (right), and is minimized when the means of $\puncond{k}$ and $\pcond{k}$ are the same.
It is through $J_\mathrm{int}$ that neural network gradients are incorporated in our framework since $\pcond{k}$ depends on $\urobt{t+1:t+T}$ (and $h$). Thus, $J_\mathrm{int}$ is differentiable with respect to $\urobt{t+1:t+T}$.

\subsection{Hamilton-Jacobi Reachability-based Safety Constraints}\label{subsec:safety_constraints}
As discussed in Section~\ref{subsec:related safe planning}, we leverage \ac{HJ} reachability analysis to provide safety assurance for the robot.
\ac{HJ} reachability reasons about closed-loop controls in response to worst-case disturbances to the system. We will briefly introduce the relevant \ac{HJ} theory here, and refer the interested reader to \cite{Herbert2020,ChenTomlin2018} for a deeper overview.
Given $\xhum$ and $\xrob$, we can define a relative state $\xrel$, and let $\dot{x}_\mathrm{rel} = f_\mathrm{rel}(\xrel, \urob, \uhum)$ describe the relative dynamics.
A natural choice is to define the relative state with respect to a coordinate frame centered around the robot. 
Let $\mathcal{T}$ be the set that we would like relative system to avoid (i.e., collision states), then

\vspace{-3mm}
{\small
\begin{align*}
    &\mathcal{A}(\tau) := \lbrace  \bar{x}_\mathrm{rel} \in \mathbb{R}^n \mid \exists \uhum(\cdot), \forall \urob(\cdot), \exists s\in[\tau, 0],   \\
    &  (\xrel(\tau) = \bar{x}_\mathrm{rel})\wedge (\dot{x}_\mathrm{rel} = f_\mathrm{rel}(\xrel, \urob, \uhum)) \wedge( \xrel(s) \in \mathcal{T}) \rbrace.
\end{align*}}
\normalsize
is the set of states that if the humans followed an adversarial policy, there does not exist a robot control policy that prevents $\xrel$ from entering $\mathcal{T}$ within a time horizon $|\tau|$ (since we are propagating backwards in time, $\tau< 0$). The set $\mathcal{A}(\tau)$ is known as the \ac{BRT}, and the robot's aim is to keep $\xrel$ outside $\mathcal{A}(\tau)$ since being inside means collision is inevitable if all the humans followed an adversarial policy.

We can compute $\mathcal{A}(\tau)$ by solving the \ac{HJI} \ac{PDE} offline via dynamic programming \cite{Herbert2020} using $\mathcal{T}$ as a boundary condition.
In particular, $\mathcal{A}(\tau)$ is the zero sub-level set of the \ac{HJI} \ac{PDE} solution $V$, also known as the value function: $\mathcal{A}(\tau) = \lbrace \xrel \mid V(\tau, \xrel) \leq 0 \rbrace.$
Solving the \ac{HJI} \ac{PDE} suffers from the curse of dimensionality and one cannot directly compute $V$ for the entire multi-agent system. Instead, we employ the system decomposition technique introduced in \cite{ChenHerbertEtAl2018} to tractably combine multiple pairwise computations to reconstruct the value function for the entire multi-agent system.
We assume $\tau$ is fixed and drop the $\tau$ argument for notational simplicity.
When $V(\xrel)$ is close to zero (i.e., near safety violation), we follow the minimally interventional control strategy proposed in \cite{LeungSchmerlingEtAl2019}. 
That is, when $V(\xrel) \leq \epsilon, \epsilon > 0$, i.e., the system is close to entering $\mathcal{A}$, we activate a safety-preserving control constraint at the current timestep,
\begin{equation}\label{eq:HJ safety preserving}
\begin{split}
    &\urob \in \mathcal{U}_\mathrm{safe}^\eta, \: \text{where} \: \:
    \mathcal{U}_\mathrm{safe}^\eta \coloneqq \big\{ \urob\in\mathcal{U}_\mathrm{R} \mid \\
    & \qquad \min_{\uhum \in \mathcal{U}_\mathrm{H}} V(\xrel +  f_\mathrm{rel}(\xrel, \urob, \uhum) \Delta t) \geq -\eta \big\},
\end{split}
\end{equation}
where $\Delta t$ is the size of the planning timestep, and $\eta$ is a slack variable to ensure feasibility of the overall \ac{TO} problem, though it is penalized heavily to keep it near zero (see Section \ref{subsec:trajopt}). When $\eta=0$, the activation of \eqref{eq:HJ safety preserving} ensures that even under adversarial policies by the humans, the system will not enter $\mathcal{A}$ within $|\tau|$ seconds. 
This approach, studied in \cite{LeungSchmerlingEtAl2019} and \cite{WangLeungEtAl2020}, has been shown to allow the robot to continually optimize the planning objective while only minimally sacrificing performance to the extent necessary to stay safe.
This is in contrast to a reactive safety controller (e.g., \cite{BajcsyBansalEtAl2019}) which \emph{switches} to the optimal \ac{HJ} control thereby ignoring the planning objective and severely impacting performance as a result.

\subsection{Minimally Interfering Trajectory Optimization Problem}\label{subsec:trajopt}
At each planning step, the robot solves the following trajectory optimization problem over a time horizon $T$ in a receding horizon fashion to reach a goal state, $x_\mathrm{g}$, while minimizing its invasiveness to the humans' behavior. For notational simplicity, the following problem is indexed relative to the current planning step,

\vspace{-3mm}
{\small
\begin{align}
    \min_{\urobt{1:T},\eta} & \: \frac{1}{T}\sum_{t=1}^{T+1} \lambda_g(\xrobt{t} - x_\mathrm{g})^2 + \lambda_\mathrm{int}J_\mathrm{int}(\urobt{1:T}; h) + \lambda_\eta \eta^2\nonumber \\
    \text{s.t.} \quad  & \xrobt{t+1} = f_\mathrm{R}(\xrobt{t}, \urobt{t}), \qquad  \forall t=0,..., T,  \nonumber \\
    & \urobt{1} \in \mathcal{U}_\mathrm{safe}^\eta, \qquad \qquad \qquad \quad \text{if } \: V(\xrel) \leq \epsilon,\label{eq:minimally invasive TO}\\
    & \xrobt{t} \in \mathcal{X}_\mathrm{R}, \qquad \qquad \qquad \quad \:\:\: \forall t=1,..., T+1,  \nonumber \\
    & \urobt{t} \in \mathcal{U}_\mathrm{R}, \qquad  \qquad \qquad \quad \:\:\: \forall t=1,..., T\nonumber \\
    & \eta \geq 0.  \nonumber
\end{align}
}
The stochasticity from human behaviors is introduced into \eqref{eq:minimally invasive TO} via $J_\mathrm{int}$. 
As described in Section~\ref{subsec:interactive_loss}, $J_\mathrm{int}$ depends on $\urobt{1:T}$ and the random variable $\Uhumt{1:T}$ to reason about how the robot affects the humans' future trajectory.
The cost function and associated weights $\lambda_g, \lambda_\mathrm{int}, \lambda_\eta$, in \eqref{eq:minimally invasive TO} encourage the robot to (i) make progress towards the goal state, (ii) interfere with humans as little as possible, and (iii) violate the \ac{HJ} reachability safety-preserving constraint as little as possible.
Further, \eqref{eq:minimally invasive TO} ensures that the safety, robot state and control, and dynamics constraints are satisfied. 
The slack variable $\eta$ ensures feasibility of the problem. Note that if $\urobt{1} \notin\mathcal{U}_\mathrm{safe}^0$, it does not immediately imply that the robot will collide into a human. A collision is only inevitable under adversarial (i.e., worst-case) human controls over the next $|\tau|$ seconds, which is unlikely assuming that the humans are also trying to avoid collisions.

\begin{figure*}[t]
    \centering
    \includegraphics[width=\textwidth]{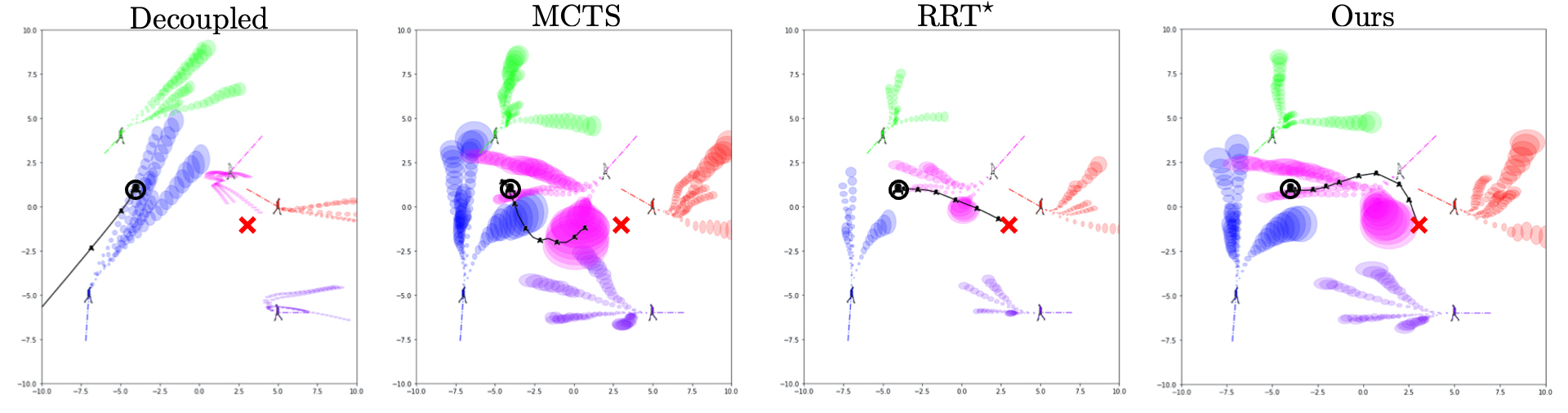}
    \caption{Qualitative comparison of different robot planning methods in a multi-agent interactive environment. Our method produces a trajectory that reaches the goal and plans around possible trajectories taken by humans. The black line represents the robot's planned trajectory, and the red cross corresponds to the robot's goal state. The colored ellipses emanating from the humans represent the robot's prediction of the humans' future trajectory conditioned on its own future trajectory and the humans' previous trajectory (dashed lines).}
    \label{fig:qualitative comparison}
\end{figure*}
\subsection{Solving the Trajectory Optimization Problem}
We assume that the dynamics and the state and control constraints are smooth, bounded, and Lipschitz continuous. The human behavior prediction model, although highly nonlinear, is differentiable and the gradients can be easily computed, e.g., via PyTorch \cite{PaszkeGrossEtAl2017}. Further, when computing the \ac{HJI} value function, we also obtain gradient information.
As such, we can solve \eqref{eq:minimally invasive TO} using a gradient-based nonlinear optimization solver. In particular, we use IPOPT \cite{WachterBiegler2006}, an interior-point solver for continuous, nonlinear, non-convex, constrained optimization problems. 
The solver returns a feasible, albeit locally-optimal, solution. 
Thus a main challenge lies in developing efficient initial guesses to \eqref{eq:minimally invasive TO}.


\subsection{Reducing Solve Time}\label{subsec:reducing solve time}
As we will discuss in Section~\ref{subsec:results_discussion}, computing $J_\mathrm{int}$ is a computational bottleneck of the proposed method.
To help reduce the solve time, we propose warm-starting strategies as well as attention mechanisms to identify which humans are relevant during each planning step. 

\noindent\emph{Warm-starting}: A good initial guess can reduce the number of solver iterations, thus computation time, required to converge to a locally-optimal feasible solution.
We investigated three warm-starting strategies:
(i) solve \eqref{eq:minimally invasive TO} but without the interactive cost term $J_\mathrm{int}$,
(ii) solve \eqref{eq:minimally invasive TO} but without the interactive cost or the \ac{HJ} control constraint, and
(iii) solve \eqref{eq:minimally invasive TO} but instead of using a neural network model for human behavior prediction, use the Social Forces model \cite{HelbingMolnar1995}, a simpler human behavior model. In general, these warm-starting strategies reduced the average optimization runtime by 5--10\% without significantly affecting the quality of the final solution. Empirically, (i) performed the best.

\noindent\emph{Attention}: We use an attention mechanism to reduce the number of agents considered when computing $J_\mathrm{int}$. This is to prevent the robot from considering interactions with humans very far away and thus not pertinent to the interaction. We investigated three attention methods: (i) all humans whose Euclidean distance from the robot is no greater than $D_\mathrm{att}$, (ii) the closest human whose Euclidean distance from the robot is no greater than $D_\mathrm{att}$, and (iii) humans who are inside the forward reachable set of the robot computed over the planning horizon.
We found that method (iii) was not very effective in reducing the number of agents to consider, while methods (i) and (ii) were able to reduce the mean computation time significantly. Depending on the number of considered agents, the computational cost can be reduced by roughly 90\% on average.
We chose (i) as it was more robust to agents moving in and out of the attention circle.


\section{Experiments}\label{sec:results}

\subsection{Experimental Set-up}\label{subsec:experimental setup}
Each human follows single integrator dynamics, and the robot follows double integrator dynamics. To reflect realistic human speed limits \cite{Bohannon1997}, the maximum speed for each human is set to 2.5ms$^{-1}$, and the maximum speed and acceleration for the robot are 2ms$^{-1}$ and 2ms$^{-2}$ respectively.
We use a planning horizon of $T=5$ seconds and a planning timestep of $\Delta t=0.4$ seconds. For the \ac{BRT} computation, we use a time horizon of $|\tau|=1$ second. Further, we use $D_\mathrm{att}=4$m to help select which human to consider when planning.

To simulate the humans in our experiments, we use another state-of-the-art human trajectory prediction model, \ac{SGAN} \cite{GuptaJohnsonEtAl2018}. \ac{SGAN} is trained on the same dataset as Trajectron++ and only conditions on previous human trajectories.
Importantly, we use \ac{SGAN}
so that all compared methods use a different prediction model than the simulator (i.e., they are not inherently advantaged).
All tests have been performed on a 2018 2.3GHz MacBook Pro. Due to the intrinsic stochasticity of the problem, all tests have been evaluated and averaged over the same 10 randomly-generated initial conditions for each configuration. Our code is located at \url{https://github.com/StanfordASL/mantrap}.


\begin{figure*}[t]
    \centering
    \subfloat[Minimum safety distance]{
    \label{fig:minimal_distance}
    \includegraphics[width=0.22\textwidth]{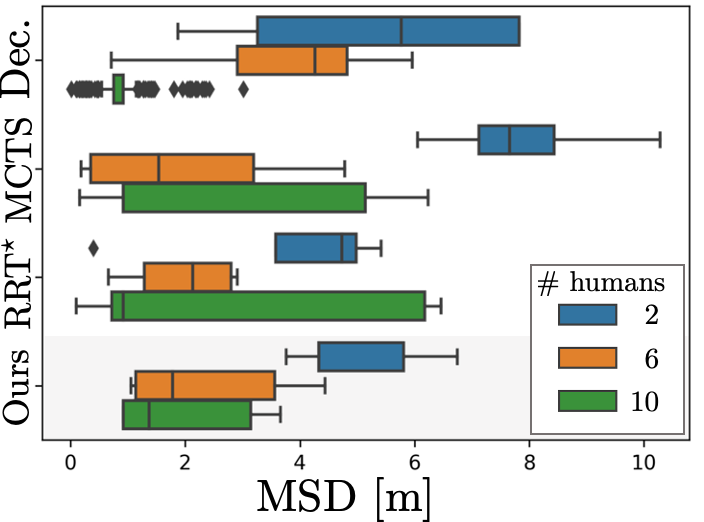}} $\:$ \subfloat[Mean robot effort as a percentage of our method's mean value.]{
    \label{fig:ego_effort}
    \includegraphics[width=0.22\textwidth]{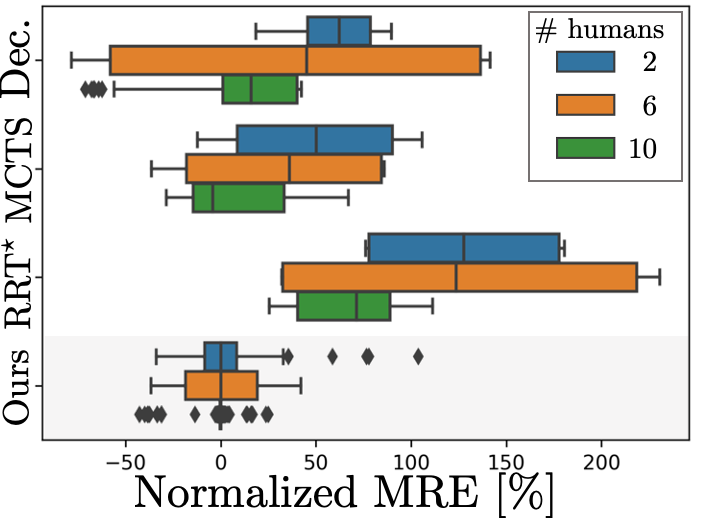}} $\:$ \subfloat[Mean pedestrian effort as a percentage of our method's mean value.]{
    \label{fig:ado_effort}
    \includegraphics[width=0.22\textwidth]{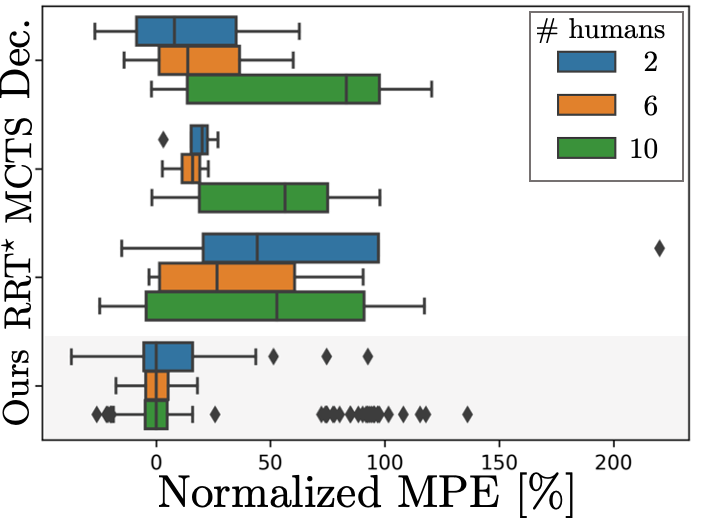}}$\:$ \subfloat[Computational time required to solve each planning step.]{
    \label{fig:runtime}
    \includegraphics[width=0.22\textwidth]{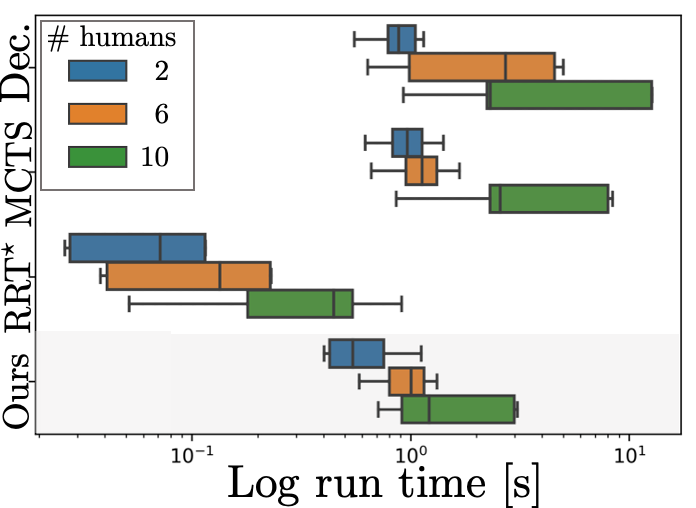}}
\caption{Our method outperforms the Decoupled, MCTS, and RRT$^\star$ baselines on the minimal separation distance (MSD), mean robot effort (MRE), and mean pedestrian effort (MPR) metrics, and is competitive in computation runtime.}
\label{fig:metrics}
\end{figure*}

\subsection{Baseline Methods}\label{subsec:baselines}
We compare against a variety of baseline approaches.

\noindent{\bf Decoupled (Dec.)}: We decouple the prediction and planning problem explicitly. Using Trajectron++, we first predict the humans' future trajectories conditioned on the robot's previously planned trajectory and hold the predictions fixed, thereby neglecting the connection between the robot and human behavior during optimization. 
The robot then solves \eqref{eq:minimally invasive TO}, but instead of the interaction loss the robot is constrained to avoid the time-dependent 1$\sigma$ covariance ellipses associated with the top five modes.


\noindent{\bf MCTS}: Monte carlo tree search is a popular policy search algorithm for decision-making in stochastic environments. Using Trajectron++, the robot simulates several scenarios of how the stochastic environment may evolve and then selects the most promising controls based on the expected cost (from \eqref{eq:minimally invasive TO}). To achieve reasonable run times, we use three Monte Carlo samples and a branching factor of three.

\noindent{\bf RRT$^\star$} \cite{KaramanFrazzoli2011}: Rapidly-exploring random tree (RRT*) is a very popular and fast sampling-based motion planning algorithm used widely in robotics. As per its standard usage, at every timestep we consider humans as static obstacles and plan a new trajectory to the goal.

\subsection{Performance Metrics}\label{subsec:performance_metrics}
To evaluate the safety and performance of our algorithm and the baselines, we use the following metrics.

\noindent{\bf \ac{MSD}}: The minimum distance experienced between the robot and all humans. To account for the trajectory between timesteps, we perform linear interpolation, indexed by $t^\prime$. Let $p$ denote position, then $\mathrm{MSD}=\min_{t^\prime, k} \| p_{\mathrm{R}, t^\prime} - p_{\mathrm{H}, t^\prime}^k \|$. A larger minimum distance implies a safer interaction (albeit with diminishing returns).
 
\noindent{\bf \ac{MRE}}: The average control effort the robot uses to execute the trajectory. $\mathrm{MRE}=\frac{1}{T}\sum_{t=1}^T \frac{\|\urobt{t}\|_2}{\|\urob{\mathrm{max}}\|_2}$. Lower values indicate more efficient  robot behaviors.

\noindent{\bf \ac{MPE}}: The $\ell_2$-norm of the difference between the mean human acceleration conditioned and unconditioned on the robot's trajectory. Let $a_\mathrm{H}^k$ denote human $k$'s acceleration sequence, then $\mathrm{MPE} = \frac{1}{TN}\sum_{k=1}^N \|\mathbb{E}_{\puncond{k}}[a_\mathrm{H}^k]- \mathbb{E}_{\pcond{k}}[a_\mathrm{H}^k]\|_2$. Lower values imply that the robot interferes less with natural human motion.






\subsection{Results and Discussion}\label{subsec:results_discussion}

We compare our method to the aforementioned baselines both qualitatively and quantitatively in multiple scenarios containing 2, 6, and 10 humans. Overall, we find that our method yields intuitive and safe behaviors (\cref{fig:qualitative comparison}) and outperforms the baselines across all metrics (\cref{fig:metrics}).           
Figure~\ref{fig:qualitative comparison} compares the different planned robot trajectories from the same initial condition. Note that the predictions differ across methods because they depend on the robot's planned trajectory.           
Due to the ellipsoidal constraints in Decoupled, the robot is prevented from moving towards the goal by the blue ellipses, leading to erratic behaviors like shooting away from the goal. 
MCTS fares better as the robot's planned trajectory heads towards the goal, however the planned trajectory is not smooth and struggles with finding a plan that avoids the prediction of a human's future trajectory (notice the robot's overlap with the pink ellipses).
RRT$^\star$ easily finds a smooth and direct trajectory to the goal since it only considers the humans as static obstacles (ignoring predictions). Like MCTS, however, RRT$^\star$ intersects with the predictions, making the planned trajectory potentially unsafe.
With our method, the robot intuitively starts by moving slowly to let the human with pink predictions move by. The robot also plans around the other pink trajectory mode corresponding to the human staying still next to the goal. In either of these outcomes, the robot has selected an intuitive plan that minimizes their interference with the humans' predicted trajectories.

For safety (see Figure~\ref{fig:minimal_distance}), our minimum MSD value is the highest compared to the other methods. Even in the most complicated scenario with 10 humans, the robot was still $\sim$1m away from the closest human whereas other methods were much closer at around 0.10m (which is a collision). Although other methods experienced larger maximum MSD values than ours, recall that the MSD metric has diminishing returns as it increases.
Figure~\ref{fig:ego_effort} shows the percent difference of each method's MRE from our method's mean MRE (lower is better). As can be seen, all other methods require a higher MRE, indicating that their robot plan is less efficient, requiring more effort to accomplish its task.
Figure~\ref{fig:ado_effort} shows the percent difference of each method's MPE from our method's mean MPE (lower is better). Overall, our method achieves the lowest MPE.
Although MCTS optimizes the same interactive cost function, it does not perform as well as our method, likely because of the limited tree search made to balance computation time. 
Since Decoupled and RRT$^\star$ do not consider $J_\mathrm{int}$ in the objective cost, these results also highlight the benefit of explicitly accounting for interaction dynamics while planning, producing nuanced behaviors and providing an additional layer of efficiency within the interaction.


Lastly, we compare the computation time required to solve each planning iteration in Figure~\ref{fig:runtime}. Unsurprisingly, RRT$^\star$ achieved the fastest computation times since it is the most lightweight and assumes a static environment. For the methods that use Trajectron++, ours has the fastest computation time and scales the best.  
In settings with six or less humans, an improved hardware and software implementation would enable our method to be deployed in real-time applications with a desired planning frequency of $2-3$Hz.
Naturally, as the number of humans in the scene increases, the optimization problem becomes larger and the runtime increases; this trend is particularly prominent for Decoupled.
While the attention mechanism described in Section \ref{subsec:reducing solve time} helps prevent the computational requirements of solving \eqref{eq:minimally invasive TO} from scaling significantly with the number of humans, the primary computational bottleneck stems from running Trajectron++. Avenues for improvement include further improving our warm-starting technique (e.g. via neural networks \cite{BanerjeeEtAl2020}), streamlining our implementation, 
and applying methods from the field of neural network compression and pruning \cite{LiebenweinBaykalEtAl2020}.


\section{Conclusions and Future Work}\label{sec:conclusions_future_work}
In this work, we present a framework for fusing together online trajectory optimization with neural network-based human behavior prediction models. The resulting robot motion planner is flexible, interpretable, and utilizes the predictive power of probabilistic, multimodal prediction models.
In particular, we focus on socially-aware navigation using Trajectron++ \cite{SalzmannIvanovicEtAl2020} 
and demonstrate that our approach produces intuitive robot behaviors that are safer and more efficient than a variety of motion planning methods.

There are three key areas of future work:
(i) improving computational efficiency with smarter warm-starting techniques, neural network compression, or using sequential convex programming with theoretical guarantees \cite{BonalliCauligiEtAl2019} (requiring less objective function evaluations compared to interior-point methods), (ii) evaluating how our method performs in more complex scenarios, such as environments with static obstacles, and (iii) validating our approach with human-in-the-loop hardware experiments and developing methods for measuring a robot's impact on human motion.


\bibliographystyle{IEEEtran}
\bibliography{IEEEabrv,ASL_papers,main}

\end{document}

%% file: notation.tex
\newcommand{\xrob}{x_\mathrm{R}} 
\newcommand{\xrobt}[1]{x_{\mathrm{R},#1}}
\newcommand{\urob}{u_\mathrm{R}} 
\newcommand{\urobt}[1]{u_{\mathrm{R},#1}}

\newcommand{\xhum}{x_\mathrm{H}} 
\newcommand{\xhumt}[1]{x_{\mathrm{H},#1}}
\newcommand{\uhum}{u_\mathrm{H}} 
\newcommand{\uhumt}[1]{u_{\mathrm{H},#1}}
\newcommand{\Uhumt}[1]{U_{\mathrm{H},#1}}

\newcommand{\xrel}{x_\mathrm{rel}}
\newcommand{\historyt}[1]{h_{#1}}

\newcommand{\puncond}[1]{p_0^{#1}} 
\newcommand{\pcond}[1]{p_\mathrm{R}^{#1}}

%% file: acronyms.tex
\begin{acronym}[Bash]
\acro{BRT}{backward reachable tube}
\acro{CVAE}{conditional variational autoencoder}
\acro{HJ}{Hamilton-Jacobi}
\acro{HJI}{Hamilton-Jacobi-Isaacs}
\acro{MPE}{Mean Pedestrian Effort}
\acro{MRE}{Mean Robot Effort}
\acro{MSD}{Minimal Seperation Distance}
\acro{PDE}{partial differential equation}
\acro{pdf}{probability density function}
\acro{POMDP}{partially observable Markov Decision Process}
\acro{RCE}{Robot Control Effort}
\acro{SGAN}{Social Generative Adversarial Network}
\acro{SQP}{Sequential Quadratic Programming}
\acro{TO}{trajectory optimization}
\end{acronym}